\documentclass[runningheads]{llncs}
\usepackage{graphicx}
\usepackage{mathtools}
\usepackage{amsmath}
\usepackage{amsfonts}
\usepackage{algorithm}
\usepackage{algorithmicx}
\usepackage{algpseudocode}
\usepackage{cite}
\usepackage{arydshln}
\usepackage{threeparttable}
\usepackage{ulem}
\usepackage{color}
\newcommand{\tblcaption}[1]{\def\@captype{table}\caption{#1}}
\begin{document}
\title{Semi-supervised Cell Detection in Time-lapse Images Using Temporal Consistency}
%
%
\author{Kazuya Nishimura\inst{1} \and
Hyeonwoo Cho\inst{1} \and
Ryoma Bise\inst{1}}


\authorrunning{K. Nishimura et al.}
%
\institute{Kyushu University, Fukuoka, Japan \\
\email{kazuya.nishimura@humna.ait.kyushu-u.ac.jp} \\
\email{bise@ait.kyushu-u.ac.jp }}
\maketitle              
\begin{abstract}
Cell detection is the task of detecting the approximate positions of cell centroids from microscopy images. 
Recently, convolutional neural network-based approaches have achieved promising performance.
However, these methods require a certain amount of annotation for each imaging condition.
This annotation is a time-consuming and labor-intensive task. 
To overcome this problem, we propose a semi-supervised cell-detection method that effectively uses a time-lapse sequence with one labeled image and the other images unlabeled.
First, we train a cell-detection network with a one-labeled image and estimate the unlabeled images with the trained network.
We then select high-confidence positions from the estimations by tracking the detected cells from the the labeled frame to those far from it.
Next, we generate pseudo-labels from the tracking results and train the network by using pseudo-labels.
We evaluated our method for seven conditions of public datasets, and we achieved the best results relative to other semi-supervised methods. Our code is available at \url{https://github.com/naivete5656/SCDTC}

\end{abstract}
\section{Introduction}
Non-invasive imaging techniques such as phase-contrast and differential interference contrast microscopy can capture images of cells without staining them.
These techniques have been widely used for the long-term monitoring of living cells, in which hundreds of cells are captured as time-lapse images at short time intervals over days. Cell detection that detects the approximate positions of cell centroids from such microscopy images is a fundamental task for biomedical research \cite{bise2015cell, kainz2015you, cruz2013deep, xu2015stacked, nishimura2019weakly, raza2019deconvolving, sirinuku2016,kainz2015you, ren2015faster, fujita2020cell}.
Time-lapse images include hundreds of cells, so manual analysis is time consuming.
Therefore, there is significant demand for automated cell-detection tools.

Traditionally, image processing-based methods such as thresholding \cite{otsu1979threshold}, level sets \cite{tse2009combined} have been proposed for cell detection \cite{bise2015cell, cosatto2008grading, xu2016automatic, vicar2019cell, veta2013automatic}. 
Recently, convolutional neural network-based detection methods have performed well with a large amount of labeled data \cite{cruz2013deep, xu2015stacked, nishimura2019weakly, raza2019deconvolving, sirinuku2016, fujita2020cell}. 
For cell detection, heatmap-based methods have achieved promising results \cite{nishimura2019weakly, raza2019deconvolving, sirinuku2016}.
However, these methods require a certain amount of annotation for each imaging condition ({\it e.g.}, type of microscopy, type of cell, and growth conditions).
Preparing the necessary amount of annotation for each condition is a time-consuming and labor-intensive task.

\begin{figure}[t]
    \centering
    \includegraphics[width=0.9\linewidth]{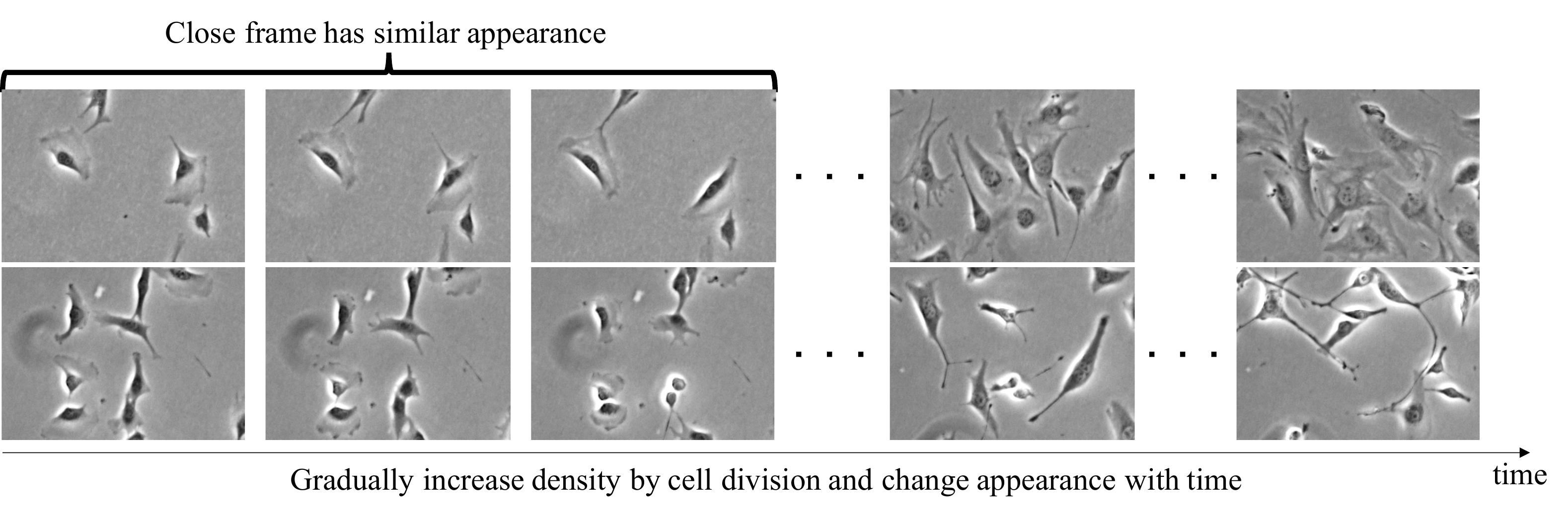}
    \caption{Appearance of sequence images. }
    \label{fig:intro}
\end{figure}

To address this annotation problem, semi-supervised object-detection methods have been proposed mainly for bounding-box-based detection \cite{Jeong2019consist, tang2021proposal, sohn2020detection, misra2015watch, wang2018cost, wang2018towards, NIPS2019_gberta, misra2015watch, kikkawa2019semi}, and a few methods have been proposed based-on consistency learning \cite{moskvyak2021semisupervised, honari2018improving}.
For example, Moskvyak {\it et al.} have proposed a consistency loss to train the network so that it consistently estimates the heatmap for data augmented by shifting or rotating \cite{moskvyak2021semisupervised}.
These consistency-based methods assume that the labeled data are randomly sampled, {\it i.e.} labeled and unlabeled data have a similar distribution.
However, this assumption often does not hold for time-lapse images of cell.
In the time-lapse images, the cell density changes due to cell division, or the characteristics of cells may change due to the cultured condition (Fig. \ref{fig:intro}). 
As a result, the appearances of cells are different between the early frames and the later frames.
Using consistency loss with an image as the label and the entire sequence as unlabeled data shows that the loss has an adversarial effect on training and does not work for the test data.

In this paper, we propose a semi-supervised cell-detection method that utilizes one time-lapse sequence, in which one image is labeled data and the other images are unlabeled data. 
Our method can improve the performance of the detection network by finding reliable detection positions from the unlabeled data using the labeled image as a clue.
First, we train a detection network $f_{d}$ with a labeled image. 
As shown in Fig. \ref{fig:intro}, if we capture time-lapse images with short intervals, cells in frames close to the labeled frame have a similar appearance to the labeled-frame cells because the appearances of cells change gradually. 
Thus, the trained network can perform an accurate estimation for these frames close to the labeled image. 
In contrast, the estimation for the later frames often does not perform well since the cell appearance changes from the labeled frame due to the increase of the density, and stimulation from culture conditions.
On the basis of this observation, we track cells using estimated results from the labeled frame to the further frames, and the consistently tracked cells are considered to be the reliable estimations. 
We use tracking results as pseudo-labels for retraining the detection CNN. By iteratively performing this process, our method can improve the detection CNN. 
Our main contributions are summarized as follows:

\begin{itemize}
    \item We propose a semi-supervised cell-detection method that can effectively train a detection network using a time-lapse sequence that contains one labeled image with the remaining images unlabeled.
    We improve the network by gradually adding pseudo-labels from the nearby frames.
    
    \item We propose a pseudo labeling method for heat map-based cell detection by using tracking. We generate pseudo-heatmaps from high-confidence detection results that selected by tracking.

    \item We demonstrated the effectiveness of our method for seven different conditions and demonstrated that our method can improve the detection network with one labeled image for various conditions.
\end{itemize}

\subsubsection{Related work:}

To address this annotation problem, semi-supervised object-detection methods have been proposed for mainly general images \cite{Jeong2019consist, tang2021proposal, sohn2020detection, misra2015watch, wang2018cost, wang2018towards, NIPS2019_gberta, honari2018improving, misra2015watch}. These methods can be divided into two groups.

The first group is consistency-based semi-supervised object detection \cite{Jeong2019consist, tang2021proposal, moskvyak2021semisupervised, honari2018improving}.
As mentioned in the introduction, the consistency-based methods assume that the labeled image is randomly sampling. 
Therefore, it does not work in our setting.

The second group is pseudo-labeling-based semi-supervised method \cite{wang2018cost, wang2018towards, wang2020co, NIPS2019_gberta, misra2015watch, li2019signet}. 
The main approach of pseudo-labeling first trains the model on a small amount of data, and samples with high confidence are selected from the estimation results. Next, the model is trained using the selected samples. The performance of the model is improved by iterative labeling and learning.
However, these methods have proposed a semi-supervised object detection method for a bounding box-based detection model. 
The cost of annotating a bounding box of a cell is expensive since a cell has a deformed shape with blurry boundaries. 
Therefore, the heatmap prediction is more suitable for cell detection tasks.
As semi-supervised learning for heatmap prediction, Gberta {\it et al.} \cite{NIPS2019_gberta} have proposed video pose propagation for sparsely annotated sequence.
By warping a heatmap from a labeled image to an unlabeled image, the method generates a pseudo label.
However, this method requires a certain pair of labeled and unlabeled images for training of the warping network, {\it i.e.}, it requires a certain labeled images in a sequence.

Unlike these methods, our method improves the detection performance with a single labeled image by gradually increasing the number of reliable pseudo-labels with tracking.

\begin{figure}[t]
    \centering
    \includegraphics[width=0.95\linewidth]{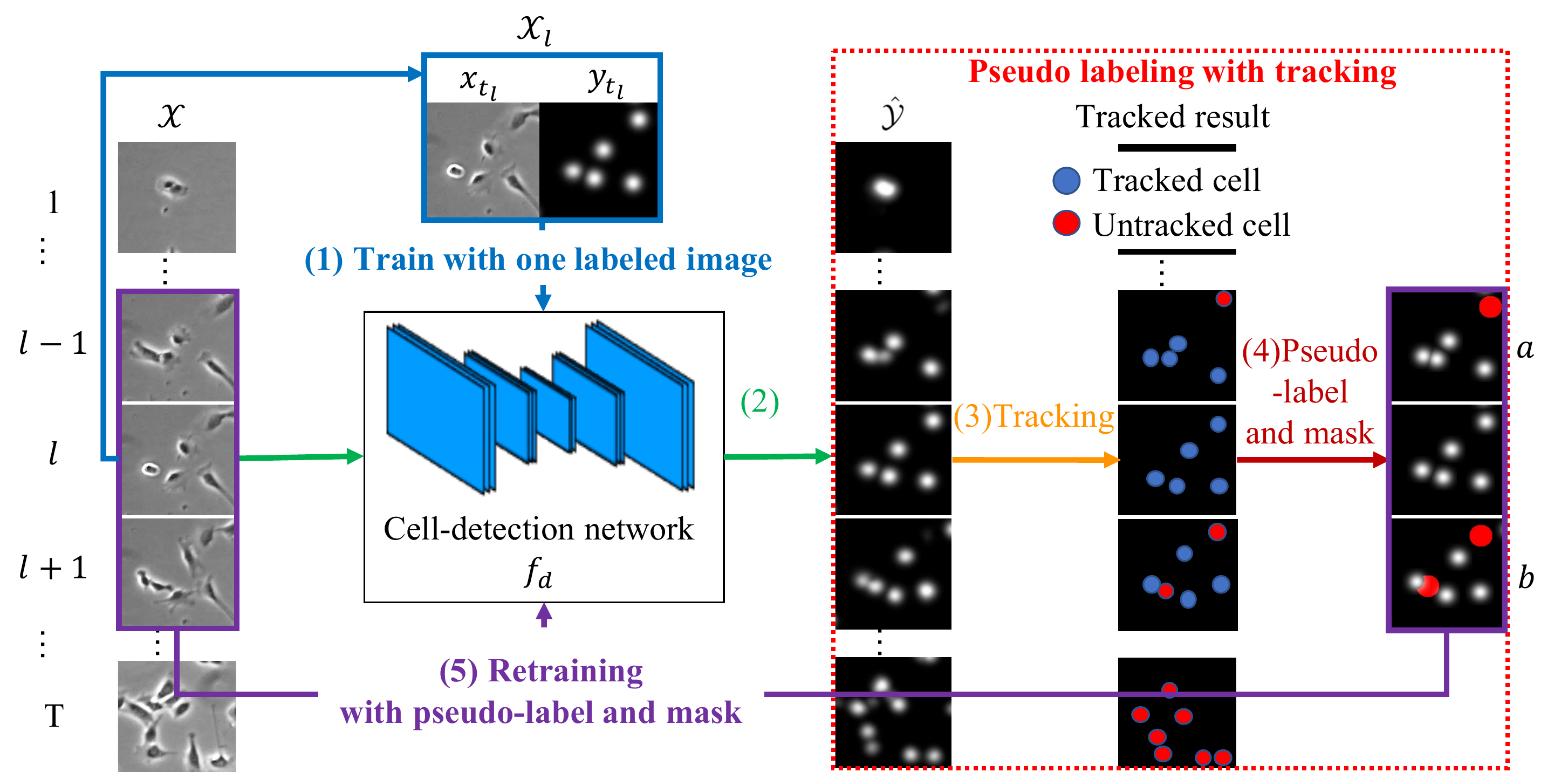}
    \caption{Overview of proposed method. First, we train a cell-detection network with one labeled image $\mathcal{X}_l$ and predict unlabeled sequence $\mathcal{X}_u$. Then, we generate pseudo-labels and masks from high-confidence detected results that is selected by the tracking. Finally, we retrain the detection CNN with selected pseudo-labels. In this process, we train the network $f_{d}$, ignoring the cell regions that are not tracked by the mask.
    The red region of the right images indicates the masked region.} 
    \label{fig:overview}
   
\end{figure}

\section{Semi-supervised cell detection}

An overview of the proposed method is shown in Fig. \ref{fig:overview}.
Given one sequence $\mathcal{X} = \{x_t \}_{t=1}^{T}$, in which there is one labeled image $\mathcal{X}_l = \{ (x_{l}, y_{l}) \}$ and the other images are unlabeled, our method improves the detection network $f_{d}$. $T$ is the number of frames in the sequence.
(1) The cell-detection network $f_{d}$ is trained with a labeled image $\mathcal{X}_l$.
(2) The prediction results $\mathcal{\hat{Y}}$ are obtained by the trained detection network $f_d$ from the whole sequence $\mathcal{X}$.
(3) The detection results are tracked from the labeled frame $l$ to the frames far from it.
(4) We generate pseudo-heatmaps $y^p_t$ and masks $ M_t $ for each frame based-on the tracking result, and we get pseudo labeled images $\mathcal{X}_p = \{(x_t, y_t^p, m_t) \}_{t=1}^{b}$. The mask $M_t$ is used on the next training step for mitigating the effect of untracked cells and mitosis cells.
(5) The network $f_{d}$ is trained with generated pseudo labels $\mathcal{X}_p$.
We improve performance by iteratively performing pseudo-labeling and learning.

\noindent
{\bf Cell detection:}
For cell detection, we use the heatmap-based detection method \cite{nishimura2019weakly}.
Given an input image $x_{t}$, the network output a heatmap $\hat{y}_t$ that is generated by blurring the approximate cell centroids.
The network is trained by the mean squared error loss between the output $\hat{y}_{l}$ and the ground-truth of the heatmap $y_{l}$.
After training, the cell position can be determined by finding the peak of the estimated heatmap.
The detected positions for each frame $\mathcal{P}= \{\Vec{p_t} \}_{t=1}^{T}$ are determined by detecting the peaks of the network output $\hat{\mathcal{Y}}=\{ \hat{y}_t \}^{T}_{t=1}$, where $\hat{y}_t = f_d(x_t)$.

\noindent
{\bf Pseudo-labeling with tracking:}
Next, we select high-confidence positions from detected positions $\mathcal{P}$ based on tracking, and generate pseudo-heatmaps using the selected high-confidence positions.
Our key assumption is that reliable estimations can be found by tracking. 
If detected positions in unlabeled frames can be continuously tracked from the labeled frame, we consider these as reliable estimations.
The other assumption is that the additional training using reliable pseudo-labels can improve $f_{d}$, and the reliable estimations increase in the next iteration by the re-trained network, which can be used as the additional pseudo-labels for further re-training. 
The network gradually improves by iterating this process.

The detection points of successive frames are associated by using a one-by-one matching \cite{KanadeT11}, which optimizes the assignment among detected points in successive frames. 
The association is performed bi-directions, from the labeled frame to far frames.
To avoid selecting the unconfident results, if the distance between the associated positions is small enough, we associate these.
The right images in Fig. \ref{fig:overview} show the examples of the estimated heat-maps $\hat{\mathcal{Y}}$ and the tracking results.
We can observe that the heat-maps of frames close to the frame $l$, were more accurately estimated compared to the results of later frames. 
Thus, the number of successfully tracked cells from $l$ gradually decreases with far from $l$. 
If the pseudo-heatmaps contain many unreliable regions, which may be cell regions or background, these affect re-training.
Therefore, we generate pseudo-heatmaps at the frames that the ratio of the tracked positions is larger than a threshold $\alpha$.
The range of the tracked frames is defined as $[a, b]$.

Next, we generate the pseudo-heatmaps from the tracking results in $[a, b]$.
We generate the set of the pseudo-heat-maps $\{y_t^p\}_{t=a}^{b}$ using the positions of the tracked cells $\{ \vec{p}_t^{tr} \}_{t=a}^{b}$ in the same manner to the supervised heatmap generation \cite{nishimura2019weakly}.
These frames still contain few unreliable regions, {\it i.e.}, the regions on un-tracked cells.
To mitigate the affect from the un-tracked cell regions, we re-train the detection network using the masked loss that ignores the masked regions in training.
There are two types of unreliable regions.

The first is a region that is detected but is not tracked. Most of this type's region is the region of a daughter cell that newly appears by cell mitosis after frame $l$. Since cells monotonically increase with time by mitosis, we only consider this type of unreliable region at $t>l$. These regions can be easily defined using an unassociated detected positions $\{\Vec{p}^{o}_t\}$. 
We define this region as $R(\Vec{p}_{t}^{o})$ that is a set of pixels within radius $\beta$ from $\{\Vec{p}^{o}_t\}$.

The second is a region that is not detected but there is a cell (miss detection).
If detected points in a certain region are continuously tracked from frame $l$ until the previous frame $t_{ut}$ {\it i.e.}, it is possible that miss-detection occurred at $t_{ut}+1$.
To define the second regions, we use the position and timing when a track was terminated based on the tracking results, in which a terminate point and the frame are denoted as $\{\Vec{p}^{ut}_{t^{ut}}\}$.
Since cells move slowly, the positions of the un-tracked cell in the later/earlier frames can be roughly predicted by random work from the terminated position and time.
A region at $t$ is defined as $R(\Vec{p}^{ut}_{t^{ut}},t)$ that is a set of pixels within radius $\beta + ||t^{ut}-t||$. The mask is defined using these unreliable regions as follows:
\begin{align}
    M_t(\Vec{p}) = 
    \begin{cases}
        0 & if ~ (\Vec{p} \in R(\Vec{p}_{t}^{ut}, t) \\ 
        0 & if ~ (\Vec{p} \in R(\Vec{p}_{t}^{o})and(t > l)\\ 
        1 & otherwise.
    \end{cases}
\end{align}
We train $f_{d}$ with $\mathcal{X}_p = \{ (x_t, y_t, M_t) \}_{t=a}^{b}$.
When we train $f_{d}$ with $\mathcal{X}_p$, we use following loss function:
\begin{align}
    L =\frac{1}{N} \sum_{N} \left(\frac{1}{\sum M_t} M_t * (y^{p}_t - \hat{y}_t) ^ 2 \right),
\end{align}
where $\Vec{i}$ is a coordinate, and $N$ is the number of pseudo-labels. We repeat 
Pseudo-labeling and re-training the network are iteratively performed until we reach $\gamma$ iterations.

\section{Experiments}
\noindent
{\bf Implementation  details:}
The detection network is trained for 10000 iterations using the Adam optimizer with a learning rate of 0.001 in each step. 
We set the hyperparameter $\alpha$, which is the error rate for terminating tracking, to $0.8$,  $\beta$, which is the radius of the region, to $18$ or $27$ according to cell size, and $\gamma$, which is the number of iterations of our pseudo-labelings, to $3$.

\noindent
{\bf Metrics:}
To evaluate detection performance, we used F1-score $= \frac{2 \cdot Precision \cdot Recall}{Recall + Precision}$, in which TP, FP, and FN are true positive,
false positive, and false negative, respectively (Precision$=\frac{TP}{TP + FP}$, Recall$= \frac{TP}{TP+FN}$).
We associate detected positions with ground-truth positions.
We define true positive as the number of associated positions. We define the number of unassociated detected positions and ground-truth positions as false positive, and false-negative, respectively.

\begin{figure}[t]
    \centering
    \includegraphics[width=0.9\linewidth]{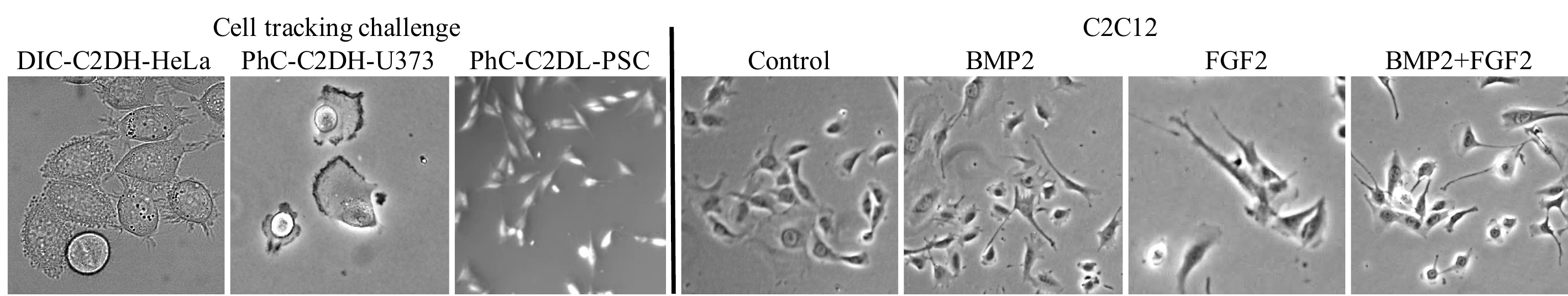}
    \caption{Example of image on each dataset.}
    \label{fig:datasets}
\end{figure}

\begin{figure}[t]
    \centering
    \begin{tabular}{c}
      \makeatletter
        \def\@captype{table}
      \makeatother
      \begin{minipage}{0.58\textwidth}
      \tblcaption{Quantitative evaluation results for Cell Tracking Challenge datasets on F1-score.}
      \centering
        \label{tab:ctc}
        \begin{tabular}{c|c| cccc} \hline
            Method & Label &HeLa & U373 & PSC & Ave. \\ \hline \hline
            Vicar \cite{vicar2019cell} &U& 0.597& 0.387&0.597 & 0.527 \\
            Baseline \cite{nishimura2019weakly} &O&0.748 & 0.879 & 0.933 &0.853 \\
            Moskvyak \cite{moskvyak2021semisupervised}&O&0.266 & 0.475 &0.820& 0.520 \\
            Ours &O& \bf{0.778} & \bf{0.914} & \bf{0.948}&\bf{0.881} \\ \hdashline 
            Moskvyak half \cite{moskvyak2021semisupervised} &H&0.714& 0.910 &0.834&0.819 \\
            Supervised \cite{nishimura2019weakly} &F&0.908 &0.925&0.972&0.935 \\ \hline 
        \end{tabular}
        \begin{tablenotes}
            \item HeLa: DIC-C2DH-HeLa, U373: PhC-C2DH-U373, PSC: PhC-C2DL-PSC, U: unsupervised, O: use one labeled image, H: use 50 \% labeled images for the first half of the sequence, F: fully supervised label.
        \end{tablenotes}
      \end{minipage}
      \begin{minipage}{0.02\textwidth}
        \hfill
      \end{minipage}
      
      \makeatletter
        \def\@captype{figure}
      \makeatother
      
      \begin{minipage}{0.45\textwidth}
      \centering
        \includegraphics[width=0.9\linewidth]{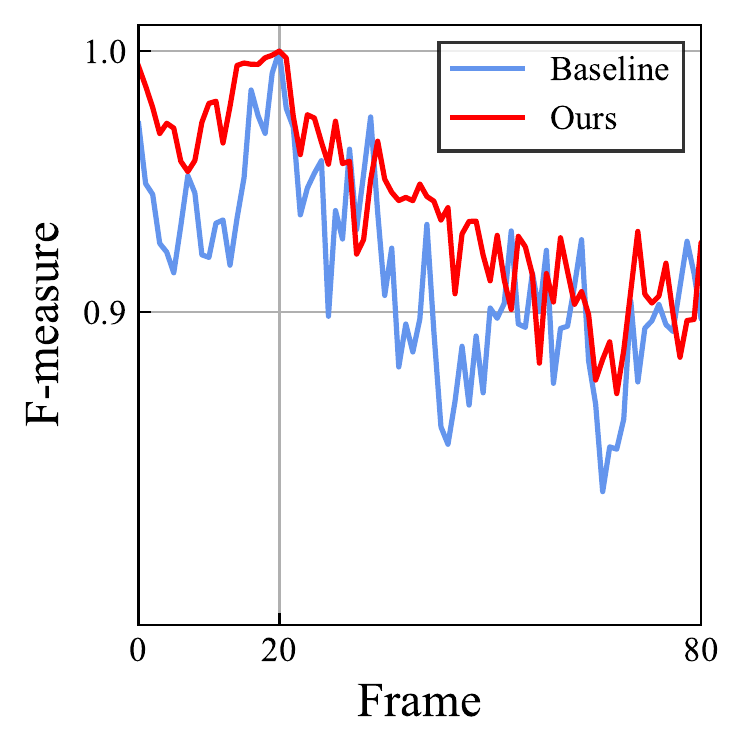}
        \caption{F1 score for each frame of training data on cell tracking challenge.}
        \label{fig:trainning}
      \end{minipage}%
    \end{tabular}
\end{figure}

\noindent
{\bf Evaluation on Cell Tracking Challenge datasets:}
To evaluate our method, we used Cell Tracking Challenge \cite{redmon2016you, ulman2017objective}, which are well-known cell image datasets.  
In this dataset, the cell were captured in various conditions as time-lapse images.
We use three type of conditions: DIC-C2DH-HeLa, PhC-C2DH-U373, and PhC-C2DL-PSC.
Two sequences that include 83 to 113 frames were fully annotated at a resolution ranging from 512x512 to 720x576 pixels. 
Since the magnification of the images (the size of cells) are different, we resized the images.
The examples of the images are shown in Fig. \ref{fig:datasets}
Please to refer to \cite{redmon2016you, ulman2017objective} for the detailed information of this dataset.
We used one image at the 20th frame as labeled frames and the rest frames as unlabeled sequence on all datasets, and we performed two fold cross validation.
We used the twentieth frame as the labeled frame and the other frames as unlabeled sequences on all datasets, and we performed twofold cross-validation.

We compared our method with 3 conventional methods using one labeled image or unlabeled: Baseline, in which Nishimura's method \cite{nishimura2019weakly} was trained by one labeled image; Vicar \cite{vicar2019cell}, which is an image processing-based method by combining preconditioning of Yin \cite{yin2012understanding} and distance transform \cite{thirusittampalam2013novel}; Moskvyak \cite{moskvyak2021semisupervised}, which is a consistency based semi-supervised detection method.
In addition, to clarify how the number of the labeled data affect to the detection performance, we also compared several methods using additional labeled data: Supervised, in which Nishimura's method \cite{nishimura2019weakly} was trained by fully labeled image ; Moskvyak half, in which the first half of the sequence was additionally labeled and used for training.

\begin{figure}[t]
    \centering
        \includegraphics[width=0.9\linewidth]{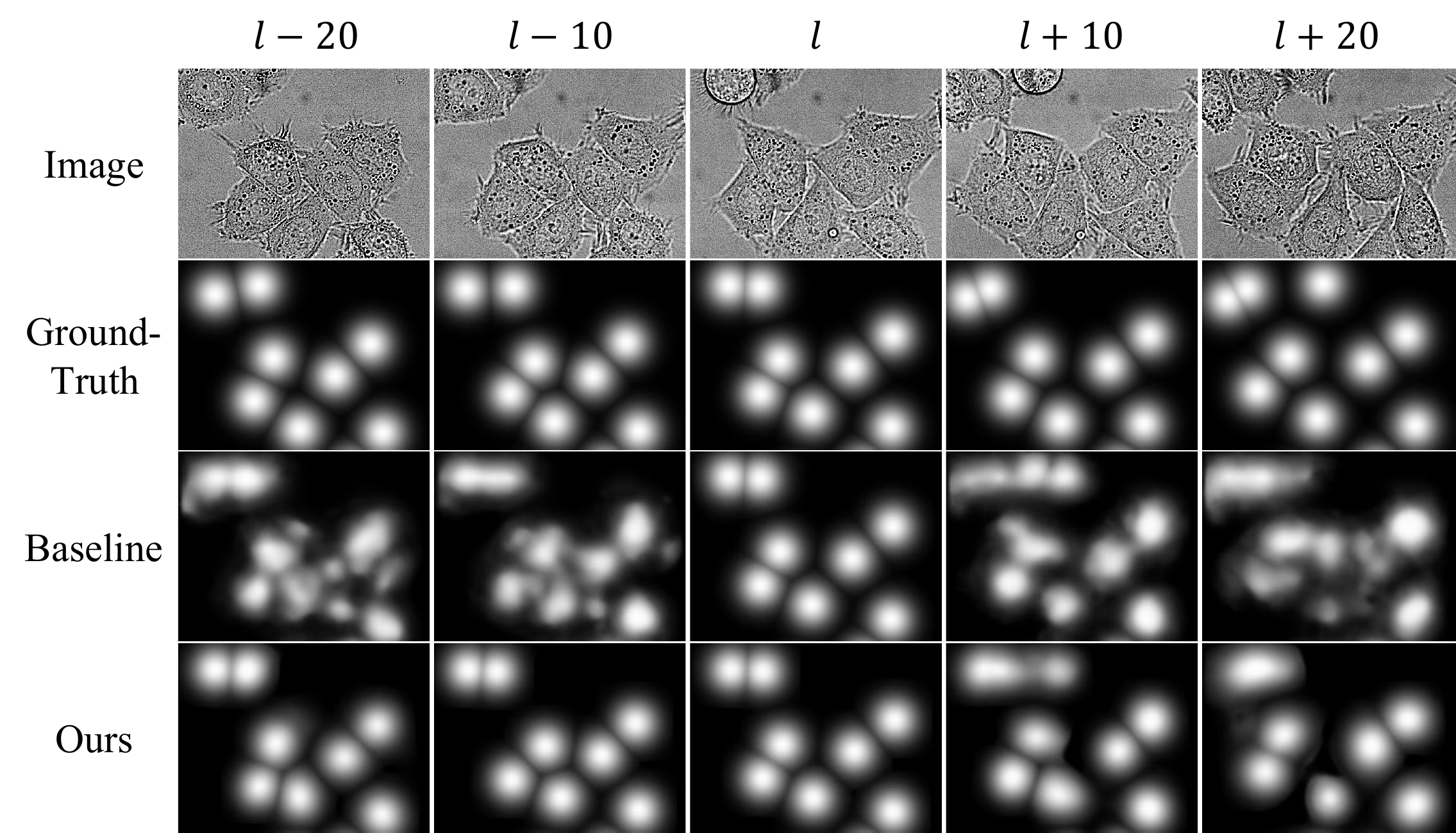}
        \caption{Example of estimation on training data.$l$ is the labeled frame.}
        \label{fig:iter_result}
\end{figure}

Tab. \ref{tab:ctc} shows a quantitative evaluation of the F1 score on Cell Tracking Challenge.
Our method achieves the best performance among the methods that trained with one labeled image.
Since the time-lapse images gradually change the appearance of the image, Moskvyak’s performance worsens performance.
Even if Moskvyak used 50 \% of labeled data, the performance decreases from the baseline on DIC-C2DH-HeLa and PhC-C2DL-PSC. 
It indicates that the previous method that assumes the labeled images are randomly sampled can not work on our target setting.
In contrast, our method can improve the performance with one labeled image.
It can observe that our method is suitable for time-lapse sequences and effectively improves performance in a semi-supervised manner.

Fig. \ref{fig:trainning} shows the average F1 score of the three datasets for each frame in the training data.
The horizontal axis indicates the F1 score, and the vertical axis indicates the frame.
The performance of the F1 score is gradually decreasing from the twentieth frame.
We observe that the performance of our method is improved compared to baseline by adding pseudo-labels from the close frames.
Fig. \ref{fig:iter_result} shows the example of result for iteration on DIC-C2DH-HeLa.
As shown in Fig. \ref{fig:iter_result}, the heatmaps of our method become more clear than the baseline even in frames away from the labeled frame.

\noindent
{\bf Evaluation on C2C12:}
To further evaluate our method on a more challenging case, we used a public dataset \cite{Ker2018phase} (C2C12), which consists of 48 sequences with 1,013 images respectively. 
C2C12 cell is captured by phase-contrast microscopy under four different media conditions (Control, FGF2, BMP2, FGF2+BMP2) at a resolution of 1040 $\times$ 1392 pixels.
An example of the images is shown in Fig. \ref{fig:datasets}. 
The appearance of the image changed depend on cultured media conditions.
Since only one sequence is fully annotated (BMP2), we additionally annotated three conditions (FGF2, Control, and BMP2 and BMP2+FGF2) to evaluate our method on various media conditions. 
We annotated 100 frames for the test data between the 600th and 700th frames: the total number of cells of is 27723, 85518, 7764, and 15082 for Control, FGF2, BMP2, and BMP2+FGF2, respectively.
We annotated the 400th frame for all conditions to a different sequence of test data as training data: the total number of cells are 116, 28, 99, and 99 for respective conditions.
Because cell type changes rapidly on FGF2 in the latter frames, we annotated frames 300 to 400 for test data and the 100th frame as training data in FGF2.
Our model is trained on a sequence that includes 1 labeled image while the other 1,012 images are unlabeled. 
We compared our method with semi-supervised and unsupervised methods that are used in previous experiments.

Table \ref{tab:quantitative_result} shows the quantitative evaluation results with the F1 score on four cultured conditions.
Our method achieves the best performance on whole-culture conditions.
Even if the cell shape slightly changes depending on the culture condition, the detection performance can be improved by this method.
It indicate that the proposed method is effective for various time-lapse images.

\begin{table}[t]
\caption{Quantitative evaluation results for C2C12 on F1-score}
\label{tab:quantitative_result}
\centering
    \begin{tabular}{l|ccccc }\hline
    Method &Control & BMP2 & FGF2  & BMP2+FGF2 & Ave.\\  \hline \hline
    Vicar \cite{vicar2019cell} & 0.731 & 0.607 & 0.676 & 0.820 & 0.709\\
    Baseline \cite{nishimura2019weakly} &0.740 & 0.844 & 0.651 & 0.824& 0.765 \\
    Moskvyak \cite{moskvyak2021semisupervised} & 0.598 & 0.170 & 0.478 & 0.520 & 0.442  \\
    Ours & \bf{0.756} & \bf{0.886} & \bf{0.830} & \bf{0.901} & \bf{0.843} \\ \hline
    \end{tabular}
\end{table}

\section{Conclusion}
We proposed a semi-supervised cell detection method for time-lapse images in which there is one labeled image and the other images are unlabeled. 
Our method can improve detection network by adding pseudo labels that is selected by tracking from detection results.
We demonstrated our method's effectiveness on seven different conditions, and we demonstrated that our method can improve detection performance on various conditions.

\noindent
{\bf Acknowledgment:}
This work was supported by JSPS KAKENHI Grant Number JP20H04211.
%
%
%
%
\bibliographystyle{splncs04}
\bibliography{refs}

\end{document}